\documentclass[sigconf,nonacm]{acmart}
\definecolor{ao(english)}{rgb}{0.0, 0.5, 0.0}
\definecolor{darkpastelgreen}{rgb}{0.01, 0.75, 0.24}

\begin{document}

\title{A Multi-Task, Multi-Modal Approach for Predicting Categorical and Dimensional Emotions}

\author{Alex-Răzvan Ispas}
\authornote{\color{darkpastelgreen} Upon reviewing the fully released version of \cite{wagner2023dawn}, it came to our attention that the proceedings version of this work did not include the speaker-independent results for \cite{wang2021fine}. Therefore, we include in brackets the speaker-independent results for a fair comparison with our model. Additionally, we were unable to ascertain any details regarding speaker-independence in \cite{cai2021speech}, and therefore advise that these results be interpreted with caution.}
\orcid{0009-0002-4034-6930}
\affiliation{%
  \institution{LISN-CNRS, Paris-Saclay University}
  \city{Saclay}
  \country{France}
}
\email{alex-razvan.ispas@universite-paris-saclay.fr}

\author{Théo Deschamps-Berger}
\orcid{0000-0003-1247-4935}
\affiliation{%
  \institution{LISN-CNRS, Paris-Saclay University}
  \city{Saclay}
  \country{France}}
\email{theo.deschamps-berger@lisn.upsaclay.fr}

\author{Laurence Devillers}
\orcid{0000-0001-9894-172X}
\affiliation{%
  \institution{LISN-CNRS, Sorbonne University}
  \city{Paris}
  \country{France}
}
\email{devil@lisn.upsaclay.fr}

\renewcommand{\shortauthors}{Ispas et al.}

\begin{abstract}
 Speech emotion recognition (SER) has received a great deal of attention in recent years in the context of spontaneous conversations. While there have been notable results on datasets like the well-known corpus of naturalistic dyadic conversations, IEMOCAP, for both the case of categorical and dimensional emotions, there are few papers which try to predict both paradigms at the same time. Therefore, in this work, we aim to highlight the performance contribution of multi-task learning by proposing a multi-task, multi-modal system that predicts categorical and dimensional emotions. The results emphasise the importance of cross-regularisation between the two types of emotions. Our approach consists of a multi-task, multi-modal architecture that uses parallel feature refinement through self-attention for the feature of each modality. In order to fuse the features, our model introduces a set of learnable bridge tokens that merge the acoustic and linguistic features with the help of cross-attention. Our experiments for categorical emotions on 10-fold validation yield results comparable to the current state-of-the-art. In our configuration, our multi-task approach provides better results compared to learning each paradigm separately. On top of that, our best performing model achieves a high result for valence compared to the previous multi-task experiments. 
\end{abstract}




\begin{teaserfigure}
  \includegraphics[width=\textwidth]{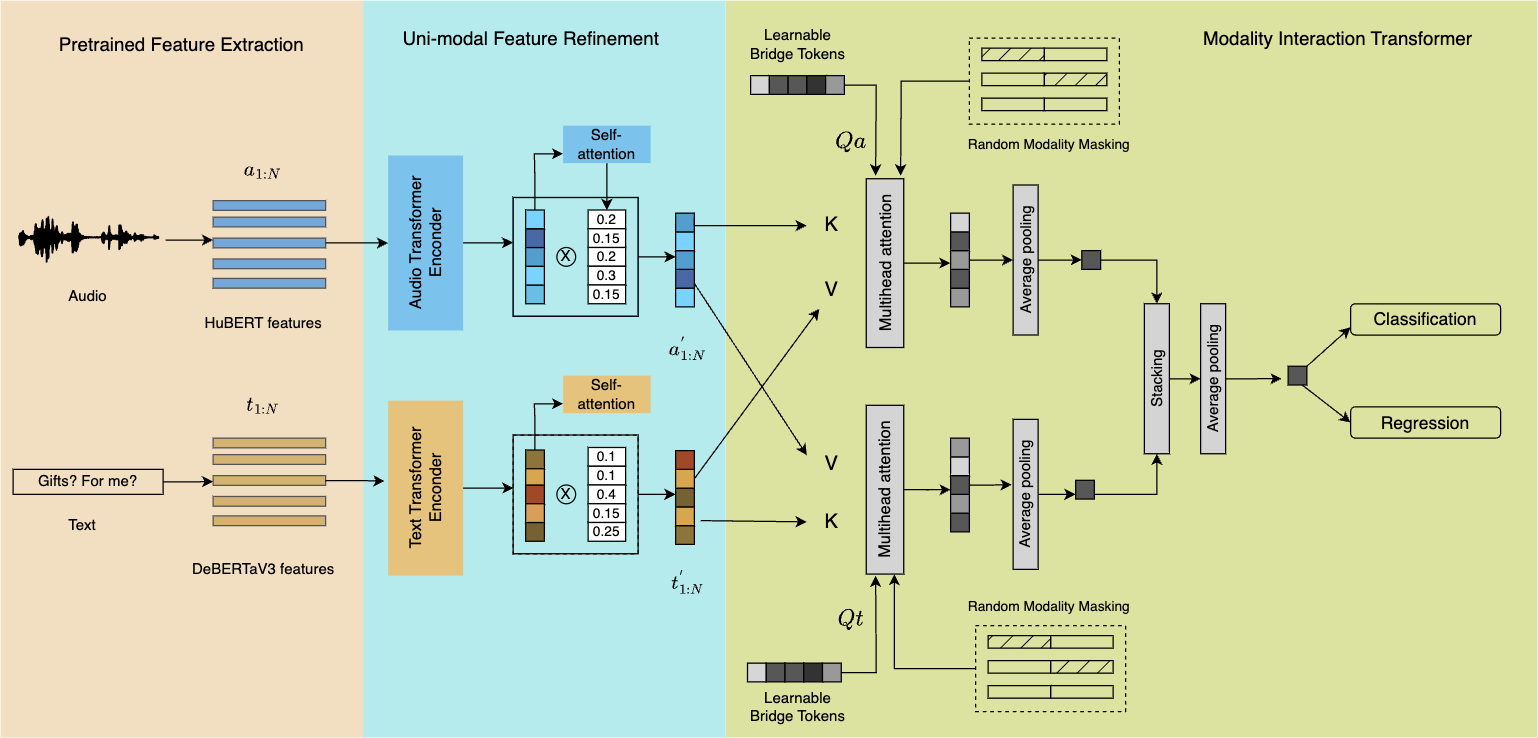}
  \caption{Overview of the proposed multi-task, cross-attention speech emotion recognition architecture. The acoustic and linguistic features are extracted with the help of HuBERT \cite{hsu2021hubert} and DeBERTaV3 \cite{he2021debertav3}. Afterwards the features are refined by two self-attention modules.  Then, inspired by \cite{deschamps2023exploring} and \cite{wang2023exploring}, we slightly modify the original modality interaction and introduce two multihead cross-attention layers which integrate the learnable bridge tokens that create the connection between the features of the two modalities. The model predicts the categorical emotion label, the valence, and the arousal score for each segment. Besides the cosine decaying procedure, the random modality masking is performed according to the original implementation.}
  \Description{The architecture is separated into three sections: the pretrained feature extraction, the uni-modal feature refinement and the modality interaction transformer.}
  \label{fig:teaser}
\end{teaserfigure}


\maketitle

\section{Introduction}

 \par The capacity to automatically estimate human users' emotions during conversations is crucial for adaptive intelligent technology. Speech is one of the most important means of communication among human beings. The paralinguistic information conveyed through speech is used to identify people's emotional states. There are two major paradigms of emotion which are referenced in the current work for affective computing. The first is the categorical representation paradigm promoted by figures like Plutchick  or Eckman \cite{ekman1992argument} who classify the emotions into groups of primary emotions (e.g \emph{surprise}, \emph{anger}, \emph{happiness}, \emph{sadness}, \emph{fear} and \emph{disgust}). The second is a continuous dimension paradigm proposed by Russel \cite{russell1977evidence} which tries to quantify emotions based on three continuous dimensions: \emph{valence}, \emph{arousal} and \emph{dominance}. 
 
 \par The IEMOCAP \cite{busso2008iemocap} dataset presents a compelling multi-modal approach to emotion analysis, combining audio and video recordings of naturalistic dyadic conversations. With diverse emotion labels representing categorical and dimensional paradigms, it offers a rich landscape for exploring and understanding human emotional expressions. Speech corpora like IEMOCAP are valuable resources which open possibilities for exploring the complex relationship between behavioral signals and emotional understanding in natural language processing and affective computing.
 
 \par While there have been successful papers which provided remarkable results for categorical emotions \cite{wang2023exploring, deschamps2023exploring, cai2021speech, aftab2022light}, as well as for dimensional emotions \cite{srinivasan2022representation, shamsi2022training}, there are few works that have tried to use both paradigms at the same time. For example, \cite{shamsi2022training}  tried to predict the categorical emotions by mapping the dimensional dimensions, meanwhile in a recent work \cite{sharma2022unifying}, the categorical and dimensional emotions are unified through a hierarchical multi-task model obtaining notable results for the arousal dimension. However, there are examples where multi-task learning proved to slightly improve the performance of other speech tasks \cite{feng2023}.

\par Multi-task learning holds promise for speech emotion recognition due to its capacity to leverage shared information between related tasks. By simultaneously optimising the model for both categorical emotion labels and continuous dimensional representations, multi-task learning enables the extraction of a more holistic and nuanced understanding of emotions. Therefore, in this paper, our aim is to demonstrate the performance contribution of multi-task learning by predicting the categorical and dimensional emotions through both acoustic and linguistic information. Considering the recent success of cross-attention between the features of the self-supervised pre-trained models of the textual and audio representations \cite{deschamps2023exploring, wang2023exploring}, we utilise a similar approach and obtain better performance on dimensional representations. We propose a multi-task architecture that predicts the categorical and the dimensional emotions at the same time, but independently. As we want to give equal importance to the acoustic and linguistic features, we adopt a similar strategy emphasised by \cite{wang2023exploring} which introduces a set of learnable query tokens which are meant to collect the emotional information from the concatenated features of the audio and textual modality. Unlike the original approach, the tokens will play the role of a \emph{bridge} that will fuse the acoustic and linguistic features together.  At the same time, we experiment with the Random Modality Masking (RMM) which masks either the audio or the text embeddings during the training in order to constrain the model to learn in depth the features of each modality.

\par Our experiments demonstrate the efficiency of cross-regularisation through categorical and dimensional multi-task learning by comparing our results with the experiments where we optimize for a single task. In addition, our best performing model achieves remarkable results for the valence dimension.  

\section{Related Work}

\par In the case of IEMOCAP \cite{busso2008iemocap}, there has been a lack of agreement regarding the number of folds as well as the number of emotional segments, making the reproducibility of the results a challenging task. Therefore, \cite{antoniou2023designing} made a review of the current IEMOCAP papers, that provided open-source implementations, in order to check their reproducibility. They also provided a set of recommendations and evaluation guidelines in which they state that a minimally comparable baseline should use the \emph{neutral}, \emph{sad}, \emph{angry} and \emph{happy+excited} classes, resulting into a total number of 5531 samples. They also recommend the use of 10-fold cross-validation where 8 speakers are used for training, 1 speaker for validation and 1 for test. Once these results are reported, the researches can modify the experiments as desired. 

\par In the context of categorical emotions, the current state-of-the-art is dominated by transformer based architectures. In the case of IEMOCAP \cite{busso2008iemocap}, fine-tuning approaches on HuBERT-large \cite{hsu2021hubert} and wav2vec2-large \cite{baevski2020wav2vec} prove to be the most efficient uni-modal methods for categorical emotions \cite{wang2021fine}. Other approaches try to use complementary tasks, such as speech recognition in order to cross-regularize the models \cite{cai2021speech}.  In multi-modal systems, cross-attention mechanisms can provide an increase of performance of 4\% \cite{deschamps2023exploring} for other categorical emotion corpora such as CEMO, an emercency call center corpus \cite{deschamps2021end}. In \cite{wang2023exploring}, they introduce \emph{random modality masking} to mitigate the issue of bias modality by maximazing the emotional information provided by both the audio and text transformer. They also introduce a group of learnable tokens in the cross-attention layer such that they can fuse the emotional features which come from the acoustic and linguistic modalities.  

\par In the case of dimensional emotions, pre-trained transformer-based architectures show better performance for valence on speech corpora, making them more robust for small perturbations and \emph{group fairness} for the two genders \cite{wagner2023dawn}.\cite{srinivasan2022representation} proposed a Teacher-Student training approach on multi-modal audio and text representations in which they achieved state-of-the-art results for dimensional values. 

\par Although there have been successful experiments that used speech-to-text in a multi-task set up to enhance the performance of categorical emotions \cite{wang2021fine, cai2021speech}, there are few papers which considered to combine the categorical and dimensional emotions. In a recent paper, both the categorical and dimensional emotions were used in a hierarchical multi-task approach \cite{sharma2022unifying}. Although the result for valence was higher, it had a slightly lower score for arousal compared to \cite{srinivasan2022representation}. \cite{chen2017multimodal} explored the prediction of multiple emotions by using multi-task learning together with every available modality (acoustic, visual and textual) in a temporal LSTM-RNN model. In another study, the arousal and valence were leveraged to assign supplementary information within a Deep Belief Network framework for the categorical task \cite{xia2015multi}.

\par In the context of our current work, we are going to respect the guidelines provided by \cite{antoniou2023designing} and compare against the other models that use the same setup. At the same time, we will follow \cite{wagner2023dawn} and use the weighted average recall (WAR) as the categorical metric since IEMOCAP is an imbalanced dataset. In the case of dimensional emotions, the metric will be the Concordance Correlation Coefficient (CCC). Our work tries to explore the possibility of performance enhancement in SER multi-modal frameworks by using both the categorical and dimensional emotions in a multi-task manner. Regarding the transformers, this work will use HuBERT-large and DeBERTaV3-large \cite{he2021debertav3}. We experimented with different combinations of the dual cross-attention emphasised in \cite{deschamps2023exploring}, the \emph{random modality masking} (RMM) and the learnable query tokens from \cite{wang2023exploring}. Our best performing model achieves a CCC score of .748 for valence.

\section{Methodology}

\par Given the audio segment and the textual transcript of someone's speech, our model is predicting the categorical emotion label, the valence and the arousal. The two transformers used to extract the acoustic and linguistic features are HuBERT-large \cite{hsu2021hubert} and DeBERTaV3-large \cite{he2021debertav3}. HuBERT-large training implies self-supervised learning in which the targets are created based on predicting clusters in which each feature vector is assigned, meanwhile DeBERTaV3-large relies on disentangled attention in order to capture dependencies between the words in the sentence. Figure \ref{fig:teaser} presents the overview of the proposed architecture. Section \ref{sec:multi-task} introduces the multi-task loss as well as its sub-components. In Section \ref{sec:refinement} we present the feature refinement as well as how we handled the sequence length issue between the audio and text embeddings. Then Section \ref{sec:modality} presents how the emotional features of the acoustic and linguistic embeddings are learned by implementing a set of learnable bridge tokens. Section \ref{sec:fusion} presents the way the complementary features are fused with the help of multi-head attention and the learnable bridge tokens. Finally, Section \ref{sec:masking} presents the Random Modality Masking (RMM) and the masking chance decaying procedure chosen alongside the training.

\subsection{Multi-task Learning}
\label{sec:multi-task}
\par Our multi-task problem can be formalised as a prediction of a set consisting of valence, arousal, and the categorical emotions which can take values from the set $\mathcal{D} = \{neutral, happy, sad, angry\}$. Given a set of features $\{u^i, t^i\}_{i=1}^I$, where $i \in \{1, \dots, I\}$ represents the i-th utterance of the speech $u^i$ and the i-th transcript  $t^i$, predict the target $y^i = [y_{Disc}^i, y_V^i, y_A^i]$, where $y^{i}$ represents the ground truth. 

\par The categorical labels will be optimised using the categorical cross-entropy (CE), which for a given dataset $I$ is computed as: 

\begin{align*}
    CE = \sum_{i=1}^I\sum_{c=1}^Dy_{i,d}\log(\hat{y}_{i,d})
\end{align*}

\par In the case of dimensional emotions, the standard loss function is the inverse of the Concordance Correlation Coefficient (CCC).  That is because the CCC score needs to be maximised, meanwhile the loss needs to be minimised. Given the vector of predictions $\hat{y} \in \mathcal{R}^I$, and the ground truth $y \in \mathcal{R}^I$, the CCC loss for one dimensional value can be calculated as: 

\begin{align*}
   \mathcal{L}_{CCC} = 1 - \frac{2Cov(\hat{y},y)}{\sigma_{\hat{y}}^2 + \sigma_{y}^2 + (\mu_{\hat{y}} - \mu_{y})} 
\end{align*}

\par Following the example of \cite{feng2023, sharma2022unifying}, we define our multi-task loss as the weighted sum of the categorical cross-entropy, the CCC loss for valence and the CCC loss for arousal, as shown in Equation 3: 

\begin{align*}
\begin{cases}
    \mathcal{L}_{total} = h_1 * CE + h_2 * \mathcal{L}_{CCC}^V + h_3 * \mathcal{L}_{CCC}^A \\
    \sum_{i=1}^{3} h_i = 1
\end{cases}
\label{eq:multi}
\end{align*}

where $h_1, h_2, h_3$ represent the hyperparameters that control the contribution of each criterion in the learning process.

\subsection{Uni-modal Refinement}
\label{sec:refinement}

\par Before performing cross-attention, we refine the encoded features with the help of the self-attention modules in order to emphasise the emotional frames for each modality. In our architecture, each self-attention module consists of 32 attention heads, which is the same amount of heads utilised for the large models in \cite{deschamps2023exploring}. Both HuBERT-large and DeBERTaV3-large have the same hidden dimension of 1024. In order to have the same representation regarding the sequence length, the shorter sequence is padded to the dimension of the longer one. The final representation of the audio and text features are $a_{1:N} \in \mathbb{R}^{Nxd_e}$ and $t_{1:N} \in \mathbb{R}^{Nxd_e}$ where $N$ is the maximum sequence length and $d_e$ is the hidden dimension size. 

\subsection{Modality Interaction}
\label{sec:modality}

\par Following the example of \cite{wang2023exploring}, we create two groups of trainable tokens which aggregate the embedding features of $a^{'}_{1:N}$ and $w^{'}_{1:N}$ by the attention mechanisms. Unlike the classic cross-attention, where the key and the value come from the same modality and the query comes from a different modality, our approach uses different modalities for the key and value and it tries to merge them together by implying a set of learnable bridge tokens. The trainable tokens $Q_a \in \mathbb{R}^{Lxd_e}$ and $Q_t \in \mathbb{R}^{Lxd_e}$ play the role of the query. Tensor $Q_a$ represents the learnable query tokens (Q) of the multihead attention layer that uses the audio embeddings as key (K) and the text embeddings as value (V), meanwhile $Q_t$ is the query for the layer where the text embeddings are the key and the audio embeddings are the value. The learnable bridge tokens can be perceived as the intermediary that provides a mean of interaction between the acoustic and linguistic features. 
Dimenson $L$ represents the number of learnable tokens. We perform cross-attention in both scenarios because it is possible that the attention weights computed with the help of either the audio or the text embeddings can offer more or less insight regarding the optimal weights for fusing the features together.
\par
\begin{figure}[h]
  \centering
  \includegraphics[width=\linewidth]{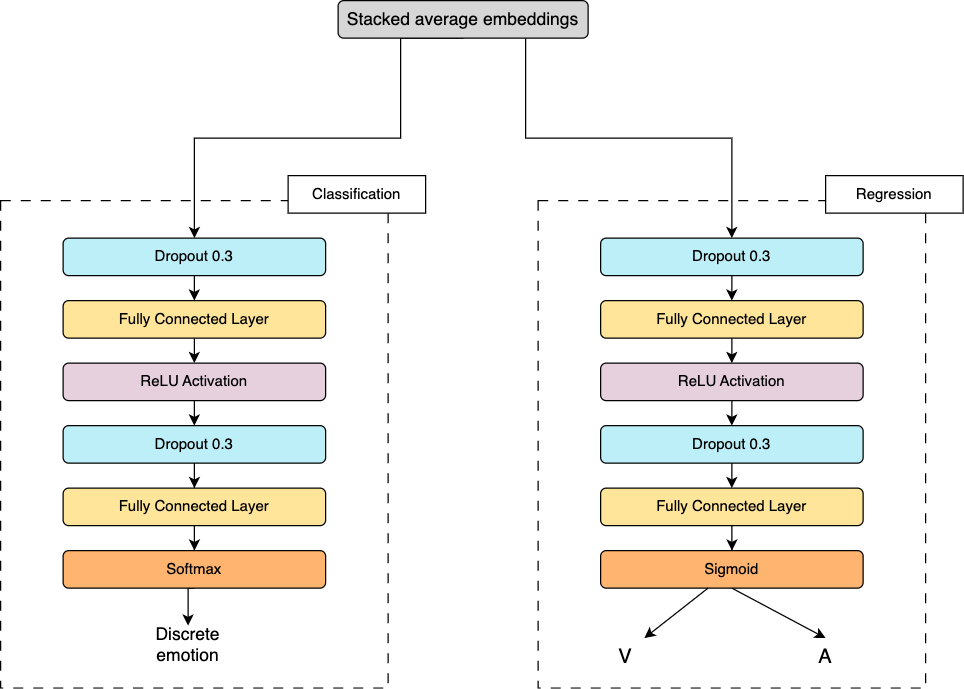}
  \caption{The head consists of a classifier and a regressor. The regressor has two sigmoid neurons which predict the valence and the arousal.}
  \label{fig:classifier_regressor}
  \Description{Both the classifier and the regressor have the same number of layers and configuration.}
\end{figure}

 Once the cross-attention is performed, each tensor will contain L tokens that are average pooled over the tokens dimension. Afterwards, the two vectors will be stacked and average pooled once again in order to get the mean emotional features from the two multi-heads. Finally, the mean vector will be fed into both the classifier and the regressor in order to predict the categorical emotion label,the valence and the arousal. The architecture of the classifier and the regressor can be seen in Figure \ref{fig:classifier_regressor}.

\subsection{Fusing the results of the multi-head cross-attention layers}
\label{sec:fusion}
\par As explained by \cite{deschamps2023exploring}, the main intuition behind attention is to weight each matrix input with respect to its importance given by the context. Cross-attention implies combining embeddings of the same dimension which come from different modalities. In the context of our architecture, instead of using a query (Q) and a key (K) from two different modalities, the queries will be learnable tensors and the key and value will come from the two modalities.

\begin{align*}
    Attention(S=(Q,K,V)) = softmax(\frac{QK^T}{\sqrt{d_k}}) V \\ 
    \text{where } S \in \{(Q_a,K_a,V_t), (Q_t, K_t, V_a)\}
\end{align*}

\par In the case when we perform cross-attention without the learnable bridge tokens, we follow the norm and we choose the key and the value to come from one modality and the query from the opposite modality.

\par As shown in Figure \ref{fig:teaser}, we followed \cite{deschamps2023exploring} and \cite{vaswani2017attention} and we implemented two multi-head attention layers such that we could apply attention mechanisms on multiple subspaces.

\begin{align*}
    Multihead(Q, K, V) = Concat(h_1, \dots, h_n) W^O \\ 
    \text{where } h_i = Attention(QW_i^Q, KW_i^K, VW_i^V)
\end{align*}

\par Once the emotional features $e_a$ and $e_t$ are obtained from the multi-head cross-attention layers, the stacked mean feature embeddings $e_s$ are computed as following: 

\begin{align*}
    e_s  = Mean(Stack(Mean(e_a),Mean(e_t))) 
\end{align*}
           
\subsection{Random Modality Masking}
\label{sec:masking}

\par As stated in \cite{wang2023exploring}, the main purpose of random modality masking (RMM) is to mitigate the situations where only the features of a single transformer are used to detect the emotions. The strategy of RMM is to encourage the model to focus on the uni-modal features, in the early stages, by randomly masking either the text or the audio embeddings. In the later stages, the model will focus more on the complementary features of the multi-modal system. In order to not alter the weights of the masked modality, when the model applies RMM we freeze the layers of the respective modality.

\par
\begin{figure}[h]
  \centering
  \includegraphics[width=\linewidth]{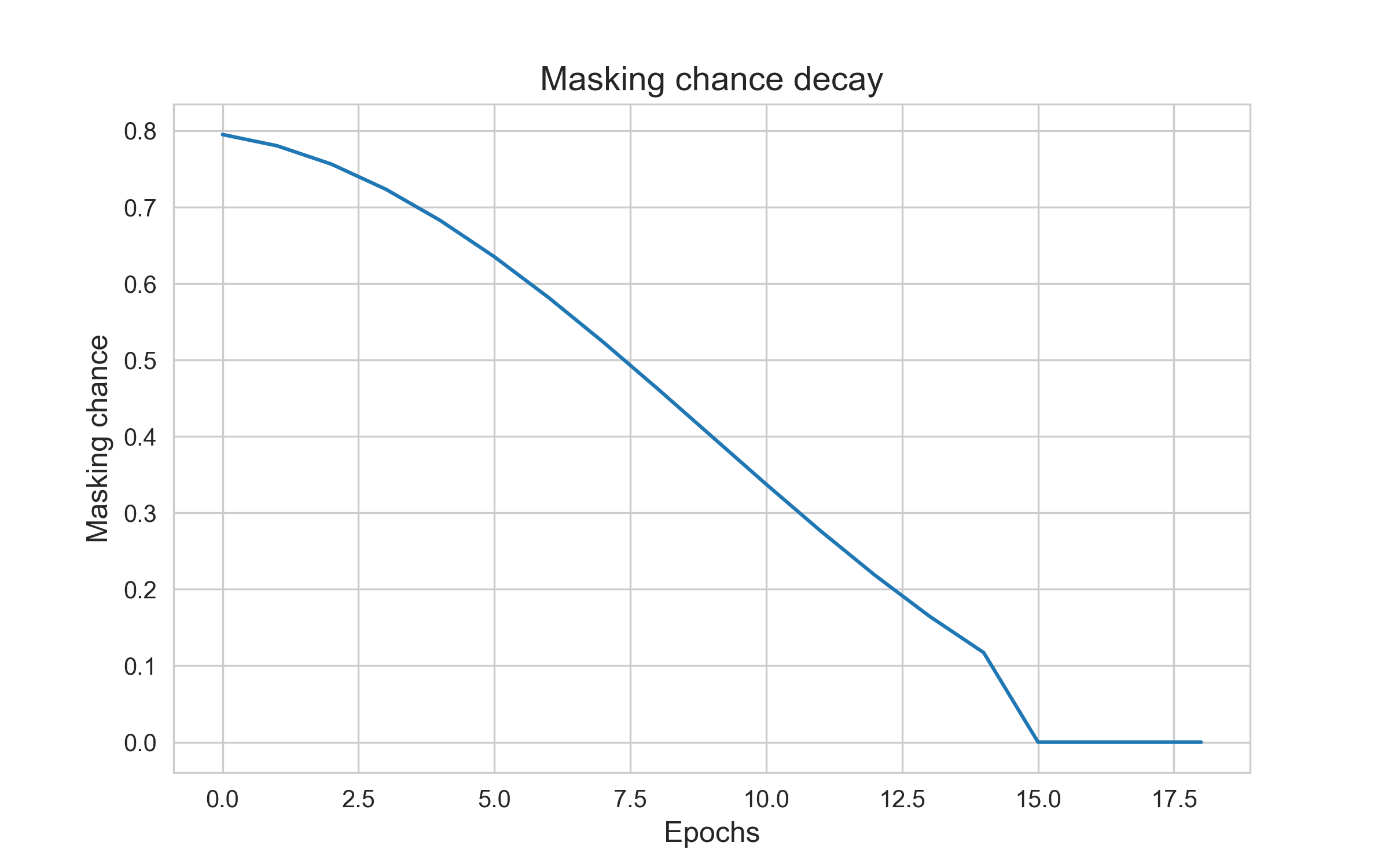}
  \caption{Masking chance cosine decay alongside the training of a fold. When the masking chance gets below 0.1, it is set by default to 0}
  \Description{Fully described in the text.}
  \label{fig:decay}
\end{figure}

 Like in the original work, we set a hyperparameter p which dictates if the respective epoch has one modality masked or not. The training starts with p set to 0.8 and it decays in a cosine manner alongside the training, as shown in Figure \ref{fig:decay}. If the probability goes below the threshold of 0.1, it is set automatically to 0. The original paper emphasised that the model is inclined to detect the emotions by relying on the features of the text transformer. Therefore, we set the RMM to be more prone to mask the textual modality. The probability of masking the text embeddings is 0.6, meanwhile the one for audio is 0.4.

\section{Experiment Design}

\subsection{Dataset}

\par The IEMOCAP dataset contains 12 hours of recorded audiovisual data, featuring five spontaneous conversational sessions between a unique male and female speaker. There are 10 speakers in total alongside 5 different sessions. Three annotators labeled each segment with categorical emotions (e.g., happy, angry, sad etc.) and provided dimensional values for the \emph{valence}, \emph{arousal}, and \emph{dominance} dimensions. The conversations are divided into scripted and improvised sections.

\subsection{Experimental Set Up}
\label{sec:set_up}

\par The experiments were performed based on the guidelines emphasised by \cite{antoniou2023designing}: we used only the \emph{neutral}, \emph{sad}, \emph{angry} and \emph{happy + surprised} labels, resulting into a total of 5531 utterances. The number of utterances for each emotion are: 1708 for \emph{neutral}, 1139 for \emph{angry}, 1084 for \emph{sad} and 1646 for \emph{happy}. In the case of dimensional emotions, we rescaled the scores from 0 to 5, into 0 to 1. At the same time, all experiments were conducted using the 10-fold validation.

\par Our experimental results are displayed in Table \ref{tab:results}. While training, we use the multi-task loss presented in Section \ref{sec:multi-task} with all hyperparameters set to 0.33 for an equal contribution. During the initial phases of our experimentation, adjustments to hyperparameters were explored, such as assigning values of 0.5 to categorical emotions and 0.25 to each dimensional variable. However, this led to a slight reduction in the performance of all tasks. Consequently, we decided to maintain the balanced weight configuration. We used the Adam \cite{kingma2014adam} optimiser, with a learning rate of $3^{-5}$ and a batch size of 16. To prevent gradient explosion in the two larger models, we implemented gradient norm clipping, setting the threshold to 1. In the experiments with learnable bridge tokens, we set the number of tokens to 30 as it was the optimal number of tokens reported in \cite{wang2023exploring}. Each multi-head attention layer had a total of 32 heads.
The embedding layers, as well as the feature extractor layers, were frozen in both modalities. Each fold was trained for 20 epochs. 

\begin{table}[h]
  \caption{Comparing Results for categorical Emotions. All experiments were conducted on 10-fold. M-T means that the respective experiment used multi-task, WAR stands for \emph{Weighted Average Recall}}
  \label{tab:commands}
  \begin{tabular}{cccl}
    \toprule
    Publication & Acoustic Features & M-T & WAR\\
    \midrule
    Feng et al. \cite{feng2020end}& MFCCs& & 68.63\%\\
    Aftab et al. \cite{aftab2022light}& MFCCs& & 70.23\%\\
    Zou et al. \cite{zou2022speech}& MFCCs, Spec, wav2vec2& & 71.64\%\\
    Wang et.al \cite{wang2021fine}& HuBERT-large & & \textbf{79.6}\%  \color{darkpastelgreen} (73.0) \\
    Wang et.al \cite{wang2021fine}& wav2vec2-large &  & 77.5\% \color{darkpastelgreen} (71.0) \\
    Cai et. al \cite{cai2021speech}& wav2vec2-large &\checkmark& 78.15\%\\
    \bottomrule

  \end{tabular}
  \label{tab:compare_categorical}
\end{table}

\begin{table*}[!h]
  \caption{Experimental results on the proposed architecture. The experiments where multi-task learning was performed have both the \emph{Disc} and \emph{Con} columns check-marked. The table specifies the components that  were used in each experiment. \emph{S/A} signifies that the model used self-attention, meanwhile \emph{CR/A} means that the model used cross-attention layers. \emph{Q} specifies if the architecture learned bridge tokens and the \emph{RMM} marks that the model is applying random modality masking during training. The metrics for categorical emotions are the \emph{Unweighted Average Recall} (UAR) and the \emph{Weighted Average Recall} (WAR). V and A represent the Concordance Correlation Coefficient (CCC) of the \emph{Valence} and \emph{Arousal}. Every experiment which implied multi-task learning was conducted on three different seeds and it displays the mean and the standard deviation.}
  \label{tab:commands}
  \begin{tabular}{cccccccccccl}
    \toprule
    Model &S/A & CR/A & Q& RMM &Disc & Con & UAR & WAR & V & A \\
    \midrule
     HuBERT-large + RoBERTa-large & & & &&\checkmark& \checkmark& 68.3 $\pm$  0.20\%&66.31 $\pm$ 0.34\%& .638 $\pm$ .01& .673 $\pm$ .03 \\
     HuBERT-large + RoBERTa-large & & & &&\checkmark& & 66.18\%&65\%& -& - \\
     HuBERT-large + RoBERTa-large & & & &&& \checkmark& - & -& .620& .674  \\     
     HuBERT-large + DeBERTav3-large & & & &&\checkmark& \checkmark& 69.43 $\pm$ 0.64\%&67.22 $\pm$ 0.74\%& .637 $\pm$ .43&.674 $\pm$ .12 \\
     HuBERT-large + DeBERTav3-large& & & &&\checkmark& & 67.12\%&65.5\%& -&- \\
     HuBERT-large + DeBERTav3-large& & & &&& \checkmark& -& -& .617&.675\\
     HuBERT-large + DeBERTav3-large& \checkmark& & &&\checkmark& \checkmark& 73.92 $\pm$ 0.17\%&73.01 $\pm$ 0.25\%& .737 $\pm$ .04&.676 $\pm$ .07\\
      HuBERT-large + DeBERTav3-large& \checkmark& & &&\checkmark& & 74.17\%&72.9\%& -&- \\
     HuBERT-large + DeBERTav3-large&\checkmark & & &&&\checkmark & -& -& .711& .687 \\
      HuBERT-large + DeBERTav3-large&\checkmark & \checkmark& && \checkmark& \checkmark& 74.68 $\pm$ 0.16\%&74.69 $\pm$ 0.18\%& .738 $\pm$ .03& .685 $\pm$ .03\\ 
     HuBERT-large + DeBERTav3-large&\checkmark & \checkmark& &&\checkmark& & 74.29\%&73.6\%& -& - \\  
      HuBERT-large + DeBERTav3-large&\checkmark & \checkmark& &&&\checkmark & -& -& .718& .686 \\ 
      HuBERT-large + DeBERTav3-large&\checkmark & \checkmark& \checkmark&&\checkmark&\checkmark & 75.71 $\pm$ 0.17\%& 74.6 $\pm$ 0.21\%& \textbf{.748} $\pm$ .05& .677 $\pm$ .04\\      
      HuBERT-large + DeBERTav3-large&\checkmark & \checkmark& \checkmark&&\checkmark& & 74.02\%& 73.05\%& -& - \\    
      HuBERT-large + DeBERTav3-large&\checkmark & \checkmark& \checkmark&&&\checkmark & -& -& .714& .698  \\
      HuBERT-large + DeBERTav3-large&\checkmark & \checkmark& \checkmark   &\checkmark&\checkmark&\checkmark & 75.79 $\pm$ 0.37\%&  74.38 $\pm$ 0.33\%& \textbf{.744} $\pm$ .02& .679 $\pm$ .04\\  
    \bottomrule
  \end{tabular}
  \label{tab:results}
\end{table*}

\section{Results}



\subsection{Comparing Results}

\par We compare our results against the other reported experiments that used 10-fold. The previous categorical results can be seen in Table \ref{tab:compare_categorical}. We gathered the same results reported by \cite{wagner2023dawn} and \cite{antoniou2023designing}. All experiments that were conducted with multi-task learning utilised speech-to-text as a secondary task. We chose the WAR metric as the primary one for categorical emotions because the dataset is imbalanced as shown in Section \ref{sec:set_up}. Therefore, we want to take into account the class frequencies and assign higher importance to the recall of the minority classes. Like that, we prevent the model from being biased towards the majority label and ensure that the performance on all four labels are considered.

\par In Table \ref{tab:compare_categorical}, the last results are obtained by using multi-task. The multi-task loss of \cite{cai2021speech} implies the categorical cross-entropy (CE)  and the Connectionist Temporal Classification (CTC) loss. In the case of the best performing model \cite{wang2021fine}, they perform partial fine-tuning on a HuBERT-large model whose output is average time pooled and fed into a linear classifier. The average time pooling is compressing the variance of the time lengths by aggregating the hidden states of the model over time. 

\par To the best of our knowledge and considering the fact that the guidelines suggested by \cite{antoniou2023designing} are recent, we could not find dimensional experiments that used 10-fold on IEMOCAP. Therefore, Table \ref{tab:compare_dimensional} contains the best performing results for dimensional emotions on 5-fold. \cite{sharma2022unifying} combined the CE together with the CCC loss of the valence, arousal and dominance.

\begin{table}[h]
  \caption{Comparing Results for dimensional Emotions. All experiments were conducted on 5-fold. M-T means that the respective experiment used multi-task. WAR stands for Weighted Average Recall}
  \label{tab:commands}
  \begin{tabular}{ccccl}
    \toprule
    Publication & Acoustic Features & M-T & V & A\\
    \midrule
    Srinivasan et al. \cite{srinivasan2022representation}& wav2vec2-base& & .363& .728\\    
    Srinivasan et al. \cite{srinivasan2022representation}& wav2vec2-large& & .472& .735\\
    Srinivasan et al. \cite{srinivasan2022representation}&HuBERT-large& & .485& .733 \\
    Srinivasan et al. \cite{srinivasan2022representation}&HuBERT-large& & .547& \textbf{.752} \\
    Roshan et al. \cite{sharma2022unifying}&& \checkmark & \textbf{.660}& .717 \\
    \bottomrule
  \end{tabular}
  \label{tab:compare_dimensional}
\end{table}

\subsection{Our Results}

\par Table \ref{tab:results} displays the results of our experiments. We chose to also display the Unweighted Average Recall (UAR) in order to emphasise the difference in performance when we do not take the sample frequency into account. If we compare the single-task and multi-task experiments from Table \ref{tab:results}, every experiment that uses both the categorical and dimensional emotions improves the CCC score of the valence. In this configuration, this result suggests that the prediction of the categorical emotions can cross-regularize the prediction of valence. This statement is also supported by the fact that the model underperforms when it is trained only for the dimensional task. Overall, in the best performing models, the multi-task improves the score of valence by 2-3\%. In the case of arousal, it was noticed that in the early epochs it was performing much better than the valence. However, in the middle stages of the training, the arousal is surpassed by the valence in almost every experiment. This behaviour could be caused by the fact that the weights of the regressor cannot be used to predict both values at the same time. The performance of the categorical emotions also benefits from multi-task learning, increasing the performance of the WAR by 1-2\% compared to the single task configuration. However, the best performing model for WAR (74.6\%) has a slightly lower performance compared to the other categorical experiments such as \cite{wang2021fine} and \cite{cai2021speech}. This result could imply that the CTC loss used by \cite{cai2021speech} is a more optimal complementary task in contrast with the regression of the dimensional emotions. In the case of \cite{wang2021fine}, unlike the standard average pooling, the average time pooling layer is considering the entire temporal dimension and it computes the average across all time steps, which seems like a more consistent method for inferring the information of the acoustic features. Overall, the multi-task learning increased the performance of every experiment, no matter the complexity of the architecture. 

\par By looking to the components of the architecture, it could be noticed that the self-attention layers play a major role in the performance, in spite of the number of optimised tasks. Our multi-task model which uses only self-attention increases the WAR of our HuBERT-large + DeBERTaV3-large baseline by almost 6\% and the CCC score of the valence by 10\%. The improvement is caused by the effectiveness of the emotional frames emphasised from each modality. Together with cross-attention, the model can reach a valence score of .738. When the learnable bridge tokens are included, the model reaches the highest score for the CCC of the valence (.748) and the second best score for the WAR (74.6\%), which suggests that the learnable bridge tokens can offer insight regarding the way of how the acoustic and linguistic features can be fused together.

\par Unfortunately, the RMM could not improve the average performance of our best model. However, the variance in performance is slightly lower across the three seeds. The underperformance of the RMM approach could be the lack of uni-modal training during the first epochs. When the RMM is applied, only the transformer whose modality features are masked gets its weights frozen. However, both multi-head attention layers are trained together. The features learned while training the two multi-head attention layers might not be beneficial for both of them. Therefore, the model could benefit more from the RMM if the multi-head attention layers are trained separately in the early stages. The classifier and regressor are also trained at the same time during all epochs. A fine-tuning epoch for both the classifier and the regressor could slightly improve the results at the end of the training. Last but not least, it has been noticed that the arousal performance downgrades in favour of the valence, which makes us consider the implementation of different regressors for each dimensional value. 

\section{Conclusion}
 
\par To conclude, we proposed a multi-task multi-modal system for predicting the categorical and dimensional emotions in spontaneous conversations. In our configuration, we compare our multi-task results with the single-task ones and emphasize that the cross-regularisation of the two tasks is beneficial for the performance of the categorical emotions and for the valence. In our setup, our approach achieves the best results by applying two groups of learnable bridge tokens that fuse the features of the two modalities through the help of the multi-head attention layers. Our best performing multi-task, multi-modal framework achieves a CCC score of .748 for valence. 

\par Unlike the previous approaches, the performance of the valence surpasses that of the arousal when the model complexity is increased. For future work, we plan to investigate the reason why the pattern gets inverted. Additionally, we intend to increase our performance by implementing the suggested improvements mentioned in the results section. We will also dive deeper into the learnable bridge tokens and find the optimal number of tokens for our configuration.

\section{Reproducibility}

\par The experiments were performed in PyTorch on two types of GPU: RTX A6000 and Tesla V100. The training of a 10-fold lasted between 16 and 24 hours depending on the number of epochs and the chosen GPU. In order to ensure reproducibility, all experiments were conducted on the same seed and were prevented from using non-deterministic algorithms. The multi-task experiments were validated by calculating the average of three distinct seeds. In order to reproduce our experiments, use the seeds 1,2 and 3 for both the CUDA and CPU versions of PyTorch. At the same time, set the PyTorch CuDNN to deterministic.

\begin{acks}
We would like to thank Emmet Strickland for their assistance in proofreading this paper. His careful review and feedback were invaluable in improving the clarity and correctness of our work.
\end{acks}


\bibliographystyle{ACM-Reference-Format}
\bibliography{sample-sigconf}


\end{document}